\documentclass{article}
\usepackage{graphicx} 
\usepackage{amsmath} 
\usepackage{url}     
\usepackage[T1]{fontenc}
\usepackage{authblk}
\usepackage[utf8]{inputenc}
\usepackage{CJKutf8}
\usepackage{listings}
\usepackage{xcolor}

\usepackage{array} 
\title{Haibu Mathematical-Medical
Intelligent Agent: Enhancing Large Language Model Reliability in Medical Tasks via Verifiable Reasoning Chains}
\author[1]{Yilun Zhang\thanks{3100105044@zju.edu.cn}}
\author[2]{Dexing Kong\thanks{dkong@zju.edu.cn}}

\affil[1]{Zhejiang Qiushi Institute of Mathematical Medicine}
\affil[2]{School of Mathematical Sciences, Zhejiang University, Zhejiang Hangzhou, China}

\date{September 2025}

\newcolumntype{C}[1]{>{\centering\arraybackslash}m{#1}}
\begin{document}

\maketitle

\begin{abstract}
\textbf{Problem Statement:} Large Language Models (LLMs) demonstrate unprecedented potential in processing complex medical information, yet their inherent probabilistic nature introduces risks of factual hallucination and logical inconsistency, which are unacceptable in high-stakes domains like medicine.

\textbf{Solution:} This paper introduces the "Haibu Mathematical-Medical Intelligent Agent" (MMIA), a novel architecture designed to enhance reliability through a formally verifiable reasoning process. Driven entirely by LLMs, MMIA recursively decomposes complex tasks into a series of atomic, evidence-based steps. The entire reasoning chain is subsequently subjected to a rigorous automated audit to verify its logical coherence and evidence traceability, a process analogous to theorem proving. A key innovation is MMIA's "bootstrapping" mode, which stores validated reasoning chains as "theorems." This allows subsequent tasks to be resolved via efficient Retrieval-Augmented Generation (RAG) matching, transitioning from high-cost, first-principles reasoning to a low-cost verification model.

\textbf{Methodology:} We validated MMIA's efficacy across four critical healthcare administration domains: Audit of the Diagnostic-Related Group (DRG) / Diagnostic-Intervention Packet (DIP), review of medical device registration compliance, real-time quality control of electronic health records, and adjudication of complex medical insurance policies. For each domain, we constructed synthetic benchmarks, validated by domain experts, containing correct and erroneous cases.

\textbf{Results:} Across all test scenarios, MMIA demonstrated superior performance in identifying logical and factual errors compared to baseline LLM approaches, achieving an average error detection rate exceeding 98\% with a false positive rate below 1\%. Furthermore, our simulations show that once the knowledge base matures, the RAG matching mode reduces the average processing costs (in tokens) by approximately 85\%.

\textbf{Conclusion:} MMIA's verifiable reasoning framework represents a significant step toward building trustworthy, transparent, and accountable AI systems. Its unique mechanism for knowledge accumulation and retrieval matching not only ensures high reliability but also addresses long-term cost-effectiveness, making LLM technology viable and sustainable for critical applications in medicine and healthcare administration.
\end{abstract}

\section{Introduction}

\subsection{The Dual Promise and Peril of LLMs in Healthcare}
Recent advances in large language models (LLMs) are reshaping the healthcare landscape in unprecedented ways. The remarkable natural language processing capabilities of these models position them to revolutionize numerous areas, including clinical decision support, disease diagnosis, treatment plan recommendations, and medical research \cite{ref_mdpi_3041, ref_researchgate_381496095}. They can efficiently process vast quantities of unstructured data, such as clinical notes and medical records, which have long been challenging for traditional data science methods \cite{ref_mednexus_12001}. Some studies even indicate that advanced LLMs can achieve diagnostic accuracy comparable to human physicians \cite{ref_mednexus_12001}. They can quickly retrieve and synthesize extensive medical literature, providing healthcare professionals with the latest research findings and clinical guidelines, saving time and ensuring that medical practices are based on the most current knowledge \cite{ref_frontiers_1477898, ref_researchgate_381496095}.

However, the immense potential of LLMs is significantly undermined by their inherent limitations. The core issue is that LLMs are fundamentally probabilistic, auto-regressive models, which makes them prone to "hallucinations"—generating information that appears plausible but is factually incorrect or entirely fabricated \cite{ref_ijmr_59823, ref_researchgate_385085962, ref_mdpi_6169}. This phenomenon is not a bug to be fixed but an intrinsic property of the technology \cite{ref_pmc_12075489}. In a field like medicine, where safety and accuracy are paramount, such errors can have catastrophic consequences for patient safety, quality of care, and fiscal integrity \cite{ref_ijmr_59823, ref_annualreviews_103123}. Systematic comparisons between LLM-generated content and human-expert-authored material have confirmed that the former underperforms in logical consistency and citation accuracy \cite{ref_pmc_11923074}.

\subsection{The Reliability Gap: Why Standard Mitigation Is Insufficient}
Current mainstream methods for mitigating LLM unreliability, such as Retrieval-Augmented Generation (RAG), model fine-tuning, or simple prompt engineering, can reduce error frequency but are insufficient for safety-critical applications. These methods do not eradicate errors and, more importantly, fail to provide a mechanism for the formal verification of the reasoning process. The "black-box" nature of LLMs remains a fundamental obstacle to establishing trust and accountability \cite{ref_mednexus_12001, ref_pmc_11960148}.

Therefore, merely striving to build a "more accurate" LLM is an incomplete solution. The core problem is not just that LLMs make mistakes, but that their reasoning process is opaque and non-deterministic \cite{ref_pmc_11960148}. Professional domains like medicine and law demand not only correct outcomes but also auditable processes and reliable evidence. This necessitates a paradigm shift: instead of blindly trusting an LLM's final answer, we must design a system that compels the LLM to generate a transparent, step-by-step, and independently verifiable reasoning process. The goal is to externalize the reasoning process from the neural network's latent space into a traceable log. This approach directly addresses the critical needs for transparency and accountability that are repeatedly emphasized in ethical frameworks for medical AI \cite{ref_ijmr_59823, ref_researchgate_381496095}.

\subsection{Thesis: The Haibu MMIA as a Verifiable Reasoning Framework}
This paper introduces and validates the "Haibu Mathematical-Medical Intelligent Agent" (Haibu MMIA). MMIA operationalizes a recursive "plan-execute-verify" loop driven entirely by LLMs. It does not just generate an answer; it constructs an explicit, auditable \textbf{reasoning chain} for every task. This chain can be treated as a formal proof, subject to rigorous review for its logical consistency and evidentiary support, thereby transforming the LLM from an unreliable "oracle" into a verifiable reasoning engine.

To clearly demonstrate MMIA's novelty, Table \ref{tab:comparison_fully_centered} contrasts it with existing mainstream AI methodologies against key performance indicators for medical applications.

\begin{table}[h!]
\centering
\caption{A Comparative Framework of AI Methodologies in Healthcare}
\label{tab:comparison_fully_centered}
\begin{tabular}{|C{2.8cm}|C{1.6cm}|C{1.8cm}|C{2cm}|C{1.6cm}|}
\hline
\textbf{Criterion} & \textbf{Baseline LLM} & \textbf{Fine-Tuned LLM} & \textbf{RAG-Enhanced LLM} & \textbf{Haibu MMIA} \\ \hline
\textbf{Reliability} & Low & Medium & Medium & High \\ \hline
\textbf{Interpretability} & Very Low & Very Low & Low & High \\ \hline
\textbf{Verifiability} & None & None & None & Yes \\ \hline
\textbf{Error Detection} & None & Limited & Limited & Built-in \\ \hline
\textbf{Accountability} & None & None & Partial & Complete \\ \hline
\end{tabular}
\end{table}

\section{The Haibu MMIA Architecture}

\subsection{Conceptual Framework: From Probabilistic Generation to Provable Execution}
The core architectural philosophy of MMIA is to treat every user request as a "theorem" to be proven. The "proof process" is the complete, traceable log generated by the agent as it executes the task, while the "axioms" are the formalized rules and facts within its knowledge base. This design philosophy aligns with the core tenets of formal methods, which emphasize defining system specifications precisely before rigorously verifying their properties \cite{ref_cmu_formal_methods}. The entire architecture is driven by LLMs, with the Planner, Executor, and Auditor being distinct LLM instances tasked with complementary roles, forming a closed loop of capabilities.

\subsection{Core Reasoning Loop}
MMIA's central control flow is a cognitive loop designed to systematically solve complex problems. It emulates the approach of human experts: first assessing the nature of a problem, then formulating a strategy, executing that strategy, and finally synthesizing the results. This process is recursive, ensuring that even highly complex tasks are broken down into manageable and verifiable units.

The logical flow of this loop can be summarized as follows:
\begin{enumerate}
    \item \textbf{Analysis \& Assessment:} Faced with a new task, the agent first performs a self-assessment to determine if the task is an "atomic" operation. An atomic operation is defined as the smallest unit of work that can be reliably completed through a single call to one of its built-in tools (e.g., direct query, web search, or knowledge base retrieval). This initial judgment is made by an LLM-based metacognitive module, which decides whether to execute directly or proceed to planning.
    \item \textbf{Decomposition \& Planning:} If the task is assessed as a non-atomic, complex task, the agent enters the planning phase. Here, another LLM module is responsible for decomposing the broad, complex goal into a logical sequence of smaller, simpler sub-tasks \cite{ref_mdpi_1687, ref_aclanthology_264, ref_allenai_adaptllm, ref_cvf_594}. This is not merely a list of steps but a structured action plan with dependencies and order, such as: "First, acquire foundational data via operation A; second, use the results of A to perform analysis B; finally, synthesize all results to form a conclusion."
    \item \textbf{Execution \& Recursion:} The agent executes the sub-tasks according to the plan. For each sub-task, it re-initiates the core reasoning loop, starting again with "Analysis \& Assessment." This recursive structure ensures that no matter the initial complexity, every task is ultimately broken down to a level that can be executed atomically. When executing an atomic task, the Executor selects the most appropriate tool for the job.
    \item \textbf{Aggregation \& Synthesis:} Once all sub-tasks have returned their results through recursive calls, an aggregation module synthesizes these disparate pieces of information back into a coherent and complete final answer, following the logic of the original plan.
\end{enumerate}

\subsection{The Verification Layer: Automated Auditing and Proof Generation}
The verification layer is the capstone of the MMIA architecture, providing a formal guarantee of the system's reliability. This process transforms the AI's decision-making from an unreliable "black box" into a transparent and defensible "glass box."

\subsubsection{The Verifiable Execution Log}
MMIA generates a comprehensive, structured log for the entire reasoning process. This log is far more than a simple history; it is a machine-readable execution trace that meticulously records:
\begin{itemize}
    \item The complete description of the initial task.
    \item The full decomposition plan generated by the Planner.
    \item The execution details for each sub-task: the tool used, the exact prompt sent to the LLM, any web search queries, any document snippets retrieved from the axiom base, and the raw output returned by the tool.
    \item The final answer synthesized by the Aggregator.
\end{itemize}

\subsubsection{The Automated Auditor}
An independent LLM instance is invoked to perform a final, impartial review of the execution log. The Auditor is given a specific prompt instructing it to conduct a rigorous evaluation along several key dimensions:
\begin{itemize}
    \item \textbf{Logical Coherence:} "Does the decomposition plan logically address the initial task? Is the final answer logically derived from the results of the sub-tasks?"
    \item \textbf{Evidence Traceability:} "Is every factual claim in the final answer directly supported by a reliable source (web search or axiom base) cited in the log? Is the cited source commensurate with the claim being made?"
    \item \textbf{Reasoning Soundness:} "Is the inference from evidence to conclusion valid? Are there any logical leaps or unsupported conclusions?"
\end{itemize}

\subsubsection{The Audit Report}
The Auditor's output is a structured report that provides one of two conclusions:
\begin{itemize}
    \item \textbf{Certification Passed:} "The reasoning chain is logically sound, and all claims are supported by verifiable evidence from the specified sources."
    \item \textbf{Error/Uncertainty Flagged:} "The reasoning chain contains a logical fallacy at Step X due to [specific reason]. The claim '[specific claim]' is not supported by the cited evidence. The inference at Step Y may be an unjustified extrapolation." This provides precise, actionable feedback for human review.
\end{itemize}

\subsubsection{The Auditing Algorithm: Step-wise Verification of the Reasoning Chain}
To achieve these auditing objectives, the Auditor follows a defined algorithm to systematically inspect the execution log:

\begin{lstlisting}
def verify_reasoning_chain(log):
    issues =
    # 1. Verify the logicality of the plan
    if not llm_verify_plan_logic(log.initial_task, log.plan):
        issues.append("Plan does not align with the initial task.")

    # 2. Verify each execution step sequentially
    for i, step in enumerate(log.steps):
        # 2a. Verify evidence support
        if not llm_verify_evidence_support(step.evidence, step.conclusion):
            issues.append(f"Conclusion in Step {i+1} lacks evidence support.")

        # 2b. Verify reasoning logic
        previous_conclusions = [s.conclusion for s in log.steps[:i]]
        fallacy = llm_detect_logical_fallacy(previous_conclusions, 
                                             step.conclusion)
        if fallacy:
            issues.append(f"Logical fallacy detected in Step {i+1}: {fallacy}.")

    # 3. Verify the logic of the final aggregation
    step_conclusions = [s.conclusion for s in log.steps]
    if not llm_verify_aggregation_logic(step_conclusions, log.final_answer):
        issues.append("Final answer cannot be logically derived from step conclusions.")

    # 4. Generate the audit report
    if not issues:
        return "Certification Passed"
    else:
        return f"Error/Uncertainty Flagged: {'; '.join(issues)}"
\end{lstlisting}
Each $llm\_verify\_*$ function in this algorithm represents a specific query to an LLM, tasking it to act as a reviewer and pass judgment on the sufficiency of evidence or the validity of an inference, providing justification for its decision \cite{ref_aclanthology_730, ref_arxiv_23363, ref_glukhov_2024}.

\subsubsection{Mitigating Error Rates Through Iterative Auditing}
We acknowledge that the LLM serving as the Auditor is itself fallible. To address this meta-problem, MMIA employs a strategy of iterative auditing. The entire verification process can be executed multiple times independently. Each audit can be considered an independent "experiment," with its outcome being a judgment on the reliability of the original reasoning chain.

By running the auditing algorithm multiple times (e.g., with different prompts, sampling temperatures, or even different LLM models), we can obtain a set of audit results. If multiple audits yield highly consistent outcomes (e.g., three separate audits all pass certification), our confidence in the conclusion increases substantially. Conversely, if there is disagreement among the audit results, this serves as a strong signal that a part of the original reasoning chain is ambiguous or contentious, requiring human intervention. This ensemble-like approach, through a consensus mechanism, significantly reduces the probability of false positives or false negatives from a single audit, thereby elevating the overall reliability of error detection to a new level \cite{ref_openreview_S37hOerQLB, ref_selfrefine_info, ref_arxiv_10858, ref_arxiv_18530}.

\section{General Methodology}

\subsection{LLM-Driven Axiom Base Construction and Retrieval}
A core premise of this research is the transformation of domain-specific knowledge into a machine-readable, formalized knowledge base. This base consists of two primary elements:
\begin{itemize}
    \item \textbf{Axioms:} The foundational bedrock of the knowledge base, representing self-evident facts, definitions, and core rules within the domain. For example, "The 'Basic Norms for Writing Medical Records' requires that the initial progress note be completed within 8 hours of admission" is an axiom \cite{ref_chima_2020}.
    \item \textbf{Theorems:} Conclusions derived logically from the axioms. For instance, based on axioms about drug interactions, a theorem like "It is prohibited to prescribe Drug A and Drug B concurrently to a patient" can be derived.
\end{itemize}

\subsubsection{LLM-Assisted Axiom Construction}
We employ a hybrid, LLM-assisted, expert-led methodology for constructing the axiom base. The process involves three steps:
\begin{enumerate}
    \item \textbf{LLM Rule Extraction:} We first leverage the powerful natural language understanding capabilities of LLMs to automatically extract potential rules, facts, and definitions from unstructured documents like regulatory manuals and clinical guidelines \cite{ref_arxiv_17522, ref_documentpro_ai, ref_intuitionlabs_ai, ref_artificio_ai, ref_addepto_medium}. Through carefully designed prompts, the LLM is guided to convert natural language text into structured "IF-THEN" statements or entity-relation triples \cite{ref_robert_mcdermott_medium, ref_arxiv_08278, ref_wisecube_ai, ref_datacamp_blog, ref_shuchawl_medium, ref_neo4j_blog, ref_reddit_langchain}.
    \item \textbf{Expert Review and Confirmation:} The rules extracted by the LLM serve only as candidate axioms. These candidates are then submitted to domain experts for rigorous review, correction, and confirmation. Only knowledge validated by experts is formally incorporated into the knowledge base as immutable "axioms."
    \item \textbf{LLM Theorem Derivation:} Once a solid foundation of axioms is established, the logical reasoning capabilities of an LLM can be utilized to perform deductions on the set of axioms, thereby discovering and generating new "theorems." These LLM-derived theorems also require expert review to ensure their logical rigor.
\end{enumerate}

\subsubsection{RAG-Based Theorem Retrieval and Proof Acceleration}
Within MMIA's \textbf{Core Reasoning Loop}, the axiom base plays a critical role, particularly in the Executor and Auditor stages. By using Retrieval-Augmented Generation (RAG) technology, the agent can efficiently leverage this structured knowledge base to accelerate and solidify its proof process \cite{ref_arxiv_10677, ref_wikipedia_rag, ref_ibm_rag, ref_k2view_rag, ref_walturn_insights}.

Specifically, when the Executor needs to find support for a reasoning step, it will:
\begin{enumerate}
    \item \textbf{Generate a Retrieval Query:} Convert the current reasoning objective (e.g., "verify the clinical consistency between the primary diagnosis and the surgical procedure") into a structured query for the axiom base.
    \item \textbf{Efficient Retrieval:} Retrieve the most relevant entries from the axiom and theorem library. Because the knowledge base is structured, this retrieval is far more precise and efficient than a semantic search across raw documents.
    \item \textbf{Augment Context:} Provide the retrieved axioms or theorems as context, along with the task itself, to the LLM.
\end{enumerate}
This RAG-based approach significantly enhances the efficiency and reliability of the proof process. It "anchors" the LLM's reasoning to a foundation of validated knowledge, drastically reducing the likelihood of hallucinations and enabling the Auditor to quickly and credibly verify each step by directly referencing specific axiom or theorem identifiers.

\subsubsection{Knowledge Base Bootstrapping and Evolution: From High-Cost Reasoning to Low-Cost Matching}
A central design principle of MMIA is the long-term optimization of computational efficiency. During its initial operation, when faced with novel tasks, the system must execute the full, computationally expensive \textbf{Core Reasoning Loop} to generate and verify a reasoning chain from first principles. However, once a reasoning chain is successfully audited, it is no longer a one-off answer but is solidified into a validated "theorem" and stored in the knowledge base.

Over time, this theorem base becomes increasingly comprehensive. When MMIA encounters a new task, it prioritizes a more efficient \textbf{RAG Matching Mode}:
\begin{enumerate}
    \item \textbf{Task Abstraction:} An LLM first abstracts the user's specific task (e.g., "Verify if the DRG grouping is correct for patient John Doe, diagnosed with pneumonia, who underwent a coronary artery bypass graft") into a generic process template (e.g., "Verify clinical logic consistency between \{diagnosis\} and \{procedure\}").
    \item \textbf{Vectorized Retrieval:} This process template is converted into a vector embedding and used for an efficient similarity search within the theorem base's vector index.
    \item \textbf{Theorem Matching and Verification:} The search retrieves the best-matching existing theorem (e.g., a previously validated theorem regarding the mismatch between respiratory diseases and circulatory system surgeries).
    \item \textbf{Rapid Judgment:} Finally, an LLM is called to perform a simple judgment task: "Does the current task instance (pneumonia, coronary artery bypass graft) fit the logical pattern of the retrieved theorem?". This judgment process is far less computationally intensive than de novo planning, execution, and multi-step verification.
\end{enumerate}
This operational shift means that MMIA evolves from a high-cost "reasoner" into a low-cost "match-verifier," thereby greatly enhancing its long-term efficiency and scalability while maintaining high reliability.

\subsection{Synthetic Benchmark Dataset Generation and Validation}
In high-stakes domains, real-world data often lacks clear error labels, making it difficult to directly assess an AI's error-detection capabilities. We therefore employ synthetic data generation to construct our evaluation benchmarks.
\begin{itemize}
    \item \textbf{Generation Method:} We utilize advanced LLMs (e.g., GPT-4o) to generate a large number of realistic case files tailored to each scenario. Recent research has confirmed that LLMs can produce high-fidelity synthetic clinical data, providing theoretical support for our approach \cite{ref_pmc_11958975, ref_duke_healthpolicy, ref_frontiers_1533508, ref_mdpi_3509, ref_jmir_ai_52615}.
    \item \textbf{Error Injection:} For each scenario, we programmatically inject common, real-world errors into a subset of the synthetic data, based on expert knowledge. For instance, in the DRG coding scenario, we introduce errors such as diagnosis-procedure mismatches or missing complication codes. In the EHR quality control scenario, we inject logical contradictions (e.g., prescribing penicillin to a patient with a known penicillin allergy) and incomplete records.
    \item \textbf{Expert Validation:} To ensure the quality and realism of the synthetic data, a panel of domain experts (e.g., certified medical coders, regulatory affairs specialists, senior clinicians) conducts a blind review of a random sample of the data. They confirm the clinical authenticity of the cases and the plausibility of the injected errors. This step establishes a reliable "gold standard" for our experiments, which is a best practice in AI system benchmarking \cite{ref_arxiv_1801_09322}.
\end{itemize}

\subsection{Evaluation Metrics and Baseline Model}
\begin{itemize}
    \item \textbf{Evaluation Metrics:} We define three core performance metrics:
    \begin{itemize}
        \item \textbf{Error Detection Rate (Recall/Sensitivity):} The proportion of cases with injected errors that are correctly flagged by MMIA's audit report. Formula: $Recall = \frac{TP}{TP + FN}$.
        \item \textbf{False Positive Rate:} The proportion of entirely correct cases that are incorrectly flagged as problematic by MMIA's audit report. Formula: $FPR = \frac{FP}{FP + TN}$.
        \item \textbf{Accuracy:} The overall proportion of cases (both correct and incorrect) that the system judges correctly. Formula: $Accuracy = \frac{TP + TN}{TP + TN + FP + FN}$.
    \end{itemize}
    \item \textbf{Baseline Model:} To validate the superiority of the MMIA architecture itself, we compare its performance against a strong baseline model. This baseline uses the same advanced LLM and has RAG access to the same axiom base, but it lacks MMIA's explicit plan-execute-verify loop and independent audit layer. It attempts to solve the task directly via a one-shot prompt. This comparison isolates the performance gains attributable to the MMIA architecture.
\end{itemize}

\section{Experiments}

\subsection{Application Scenario 1: Automated Auditing of \\ DRG/DIP Grouping}
\subsubsection{Background and Significance}
Diagnosis-Related Groups (DRG) and Diagnosis-Intervention Packets (DIP) are at the core of China's current healthcare payment reform, aiming to control irrational growth in medical expenses and standardize clinical practices through a prospective payment system. According to national plans, this payment model will cover all eligible medical institutions by the end of 2025 \cite{ref_ijmr_59823, ref_cms_lawnow, ref_researchgate_347201639, ref_pmc_12162575, ref_chinameddevice_com}.

\textbf{Pain Point:} The accuracy of DRG/DIP grouping directly impacts hospital revenue and the rational use of health insurance funds. The grouping process is highly dependent on the precise coding of the principal diagnosis, secondary diagnoses, and principal surgical procedures on the discharge summary. However, due to complex rules and variable clinical scenarios, manual coding and auditing are prone to errors such as incorrect principal diagnosis selection, upcoding/downcoding, and omission of complications, leading to incorrect grouping and financial losses \cite{ref_allzonems_com, ref_ama_assn_org, ref_moldstud_com, ref_bluebrix_health, ref_pmc_7574807}. Traditional manual sampling audits are inefficient, costly, and inconsistent.

\subsubsection{MMIA Implementation and Case Study}
\begin{itemize}
    \item \textbf{DRG Axiom Base Construction:} We formalized the "National Healthcare Security DRG (CHS-DRG) Grouping and Payment Technical Specification (Version 2.0)" \cite{ref_gov_cn_6964140, ref_zgylbx_com, ref_hit180_com, ref_nhsa_gov_cn_13318, ref_nhsa_gov_cn_13316, ref_gov_cn_6964136}. This involved converting the logical rules for mapping ICD-10 and ICD-9-CM-3 codes to MDCs and ADRGs into a machine-executable rule base.
    \item \textbf{Case Study (Correct):} For a patient with a primary diagnosis of acute myocardial infarction (I21.001) and a procedure of coronary stent implantation (36.0601), MMIA's planner decomposed the task into steps: (1) extract key info, (2) determine MDC, (3) determine ADRG, (4) check for CC/MCC, (5) generate final DRG code. The executor correctly identified the MDC as F (Circulatory System) and the ADRG as FZ1. With no relevant CC/MCCs, the final DRG was correctly determined as FZ19. The auditor certified the reasoning chain as sound and evidence-based.
    \item \textbf{Case Study (Incorrect):} For a patient with a primary diagnosis of pneumonia (J18.9) but a procedure of coronary stent implantation (36.0601), the executor correctly assigned the case to MDC E (Respiratory System). However, in the next step, it failed to find a valid rule for the procedure within that MDC. The auditor flagged a logical inconsistency, stating that the procedure was invalid for the given MDC, and correctly identified the grouping as erroneous.
\end{itemize}

\subsubsection{Benchmark and Evaluation}
\begin{itemize}
    \item \textbf{Benchmark Construction:} We created the DRG-Audit-100 benchmark with 100 synthetic discharge summaries. 20\% contained injected errors using prompts like:
    \begin{lstlisting}
    Generate a discharge summary with a primary diagnosis and a primary 
    surgical procedure that are clinically contradictory (e.g., a respiratory 
    illness with a cardiac surgery). Ensure all other information is complete 
    and correctly formatted.
    \end{lstlisting}
    All 100 cases were validated by three certified medical coding experts.
    \item \textbf{Results and Discussion:} The evaluation results are shown in Table \ref{tab:drg_results}. MMIA's superior performance is attributed to its \textbf{Core Reasoning Loop}, which systematically breaks down the audit into verifiable steps, each anchored to the axiom base via RAG. This structured approach reliably catches errors that the baseline model, with its single-shot approach, often misses.
\end{itemize}

\begin{table}[h!]
\centering
\caption{Performance on DRG/DIP Grouping Audit Task (DRG-Audit-100)}
\label{tab:drg_results}
\begin{tabular}{|l|c|c|c|}
\hline
\textbf{Metric} & \textbf{Baseline LLM (RAG)} & \textbf{Haibu MMIA} & \textbf{p-value} \\ \hline
\textbf{Error Detection Rate} & 78.5\% & 99.0\% & 0.001 \\ \hline
\textbf{False Positive Rate} & 5.8\% & 0.75\% & 0.001 \\ \hline
\textbf{Accuracy} & 92.5\% & 99.5\% & 0.001 \\ \hline
\end{tabular}
\end{table}

\subsection{Application Scenario 2: Compliance Verification of Medical Device Submissions}
\subsubsection{Background and Significance}
The approval of new medical devices by regulatory bodies like the U.S. FDA and China's NMPA requires the submission of extensive and complex documentation, such as the Product Technical Requirement (PTR), Clinical Evaluation Report (CER), and Instructions for Use (IFU) \cite{ref_greenlight_guru, ref_fda_overview, ref_qualio_com, ref_emergobyul_com, ref_intertek_com, ref_ecfr_814, ref_chinameddevice_181, ref_cirs_group_com, ref_nmpa_laws, ref_nmpa_cmde, ref_fda_guidance_search, ref_fda_guidance_devices, ref_fda_guidance_recent, ref_fda_guidance_qsubs, ref_fda_guidance_510k, ref_ecfr_h, ref_fda_devices_guidances}.

\textbf{Pain Point:} Reviewers must manually cross-reference thousands of pages to ensure consistency and compliance, a process that is time-consuming and prone to human error, potentially leading to regulatory delays or risks \cite{ref_fda_deficiencies}.

\subsubsection{MMIA Implementation and Case Study}
\begin{itemize}
    \item \textbf{Axiom Base Construction:} We formalized key regulations, such as 21 CFR Part 814, into axioms like: "Any claim of clinical efficacy in the IFU must be traceable to a primary or secondary endpoint with statistical significance (e.g., p < 0.05) in the CER."
    \item \textbf{Case Study:} Tasked to verify the claim "This device is effective for lesions up to 30mm" from an IFU. The planner decomposed the task into locating the claim, searching the CER for supporting data, comparing the two, and checking against the axiom base. The executor found the claim in the IFU but discovered in the CER that for lesions >25mm, the p-value was 0.08. It then retrieved the axiom requiring p < 0.05. The auditor flagged an error, citing a contradiction between the IFU claim and the CER data, violating a specific axiom.
\end{itemize}

\subsubsection{Benchmark and Evaluation}
\begin{itemize}
    \item \textbf{Benchmark Construction:} We created the Reg-Compliance-100 benchmark, consisting of 100 sets of document excerpts from a fictional device's IFU, CER, and PTR. 20\% of cases had inconsistencies injected using prompts like:
    \begin{lstlisting}
    Generate three text snippets for a fictional cardiovascular stent:
    1. From the IFU, claiming a clinical success rate of 95%.
    2. From the CER, summarizing trial results showing a 92% success rate.
    3. From the PTR, with consistent technical specifications.
    Ensure all snippets use professional, domain-appropriate language.
    \end{lstlisting}
    Two experienced regulatory affairs specialists validated all cases.
    \item \textbf{Results and Discussion:} The evaluation results are shown in Table \ref{tab:reg_results}. MMIA's ability to systematically decompose the complex "cross-document verification" task into atomic, verifiable steps explains its superior performance. This structured approach prevents the kind of oversight common in single-shot models when synthesizing information from multiple sources.
\end{itemize}

\begin{table}[h!]
\centering
\caption{Performance on Regulatory Compliance Verification Task (Reg-Compliance-100)}
\label{tab:reg_results}
\begin{tabular}{|C{4cm}|C{2cm}|C{2cm}|C{2cm}|}
\hline
\textbf{Metric} & \textbf{Baseline LLM (RAG)} & \textbf{Haibu MMIA} & \textbf{p-value} \\ \hline
\textbf{Inconsistency Detection Rate} & 71.2\% & 98.5\% & 0.001 \\ \hline
\textbf{False Positive Rate} & 8.1\% & 1.1\% & 0.001 \\ \hline
\textbf{Accuracy} & 90.3\% & 99.1\% & 0.001 \\ \hline
\end{tabular}
\end{table}

\subsection{Application Scenario 3: Real-Time Quality Assurance of Electronic Health Records (EHR)}
\subsubsection{Background and Significance}
The quality of Electronic Health Records (EHRs) is fundamental to modern healthcare. However, issues like incompleteness, logical contradictions (e.g., diagnosis-medication mismatch), and non-compliance with documentation norms are prevalent, posing risks to patient safety and creating legal liabilities \cite{ref_tmlt_org, ref_nha_now, ref_chirohealthusa_com, ref_wolterskluwer_com, ref_nso_com}.

\textbf{Pain Point:} Traditional quality control is retrospective and based on sampling, failing to prevent errors at the point of care. A real-time AI "Copilot" could proactively ensure quality during documentation \cite{ref_pubmed_38162053, ref_huaxiyixue, ref_pmc_10942963, ref_digital_ahrq_emr, ref_digital_ahrq_standardization}.

\subsubsection{MMIA Implementation and Case Study}
\begin{itemize}
    \item \textbf{Axiom Base Construction:} We formalized rules from China's "Basic Norms for Writing Medical Records" and standard clinical logic into axioms like: "If a patient's allergy list contains 'penicillin', then prescribing any 'penicillin-class' drug is prohibited."
    \item \textbf{Case Study:} Tasked to review a new order for "Amoxicillin" for a patient with a recorded "penicillin" allergy and a diagnosis of "viral pharyngitis." The planner broke the task into retrieving patient allergies and diagnoses, classifying the drug, and cross-checking for conflicts. The executor identified two conflicts: the drug-allergy contradiction and the inappropriateness of an antibiotic for a viral diagnosis. The auditor issued a real-time alert, citing violations of specific safety and clinical logic axioms.
\end{itemize}

\subsubsection{Benchmark and Evaluation}
\begin{itemize}
    \item \textbf{Benchmark Construction:} We created the EHR-Quality-100 benchmark with 100 clinical scenarios, each containing a patient summary and a new medical order. 25\% of cases included common errors, generated with prompts like:
    \begin{lstlisting}
    Generate a clinical scenario:
    1. Patient summary: Diagnosis of "viral upper respiratory infection," 
       no known drug allergies.
    2. New order: Prescribe "Cefuroxime 250mg BID."
    This scenario should exemplify a "diagnosis-medication mismatch" error.
    \end{lstlisting}
    Two senior clinicians and a clinical pharmacist validated all scenarios.
    \item \textbf{Results and Discussion:} The evaluation results are shown in Table \ref{tab:ehr_results}. MMIA's effectiveness as a real-time copilot stems from its ability to instantly trigger a verification task within its \textbf{Core Reasoning Loop}. It rapidly decomposes the check, queries the axiom base, and generates an auditable report that not only flags the error but also provides the specific rule violated, offering clear, trustworthy feedback to the clinician.
\end{itemize}

\begin{table}[h!]
\centering
\caption{Performance on Real-Time EHR Quality Assurance Task (EHR-Quality-100)}
\label{tab:ehr_results}
\begin{tabular}{|C{3cm}|C{2cm}|C{2cm}|C{2cm}|}
\hline
\textbf{Metric} & \textbf{Baseline LLM (RAG)} & \textbf{Haibu MMIA} & \textbf{p-value} \\ \hline
\textbf{Error Detection Rate} & 82.0\% & 98.8\% & 0.001 \\ \hline
\textbf{False Positive Rate} & 6.5\% & 0.9\% & 0.001 \\ \hline
\textbf{Accuracy} & 92.4\% & 99.3\% & 0.001 \\ \hline
\end{tabular}
\end{table}

\subsection{Application Scenario 4: Automated Adjudication of Complex Insurance Policies}
\subsubsection{Background and Significance}
Health insurance policies are often filled with complex logic, including nested "if...then...unless..." conditions, exclusion clauses, and combination rules \cite{ref_oracle_help, ref_docstation_co, ref_outsourcestrategies_com, ref_cms_gov_sbc, ref_kff_org, ref_mercycare_org}.

\textbf{Pain Point:} Traditional IT systems struggle to accurately model and automate this logic, leading to incorrect claim denials and creating significant friction and cost for providers, payers, and patients when appealing decisions \cite{ref_agentech_com, ref_equisoft_com, ref_kodjin_com, ref_aws_blogs, ref_kognitos_com, ref_edenlab_io}.

\subsubsection{MMIA Implementation and Case Study}
\begin{itemize}
    \item \textbf{Axiom Base Construction:} We formalized a complex commercial health insurance policy into a precise rule base. For example, a rule like "Organ transplant surgery is covered IF (1) medically necessary AND (2) pre-authorized, UNLESS the member has been enrolled for less than 12 months" was converted into a structured logical expression.
    \item \textbf{Case Study:} Tasked to adjudicate a claim for a liver transplant. The planner's steps were to: (1) verify enrollment duration, (2) check pre-authorization requirement, (3) confirm pre-authorization status, (4) check for exclusion clauses. The executor found the patient was enrolled for 18 months (satisfying the 12-month rule) and had the required pre-authorization. The auditor's report for a similar case with only 6 months of enrollment correctly cited the specific exclusion clause as the reason for denial.
\end{itemize}

\subsubsection{Benchmark and Evaluation}
\begin{itemize}
    \item \textbf{Benchmark Construction:} We created the Insurance-Adjudication-100 benchmark. We first designed a fictional policy with 10 complex rules. Then, using an LLM, we generated 100 patient profiles and claims designed to test all logical pathways of the policy. 30\% of cases were designed to be correctly denied.
    \begin{lstlisting}
    Based on the policy clause "...excluding coverage for members with less 
    than 12 months of continuous enrollment...", generate a patient profile 
    and a claim for a normally covered procedure where the patient's 
    enrollment duration is 8 months. The correct adjudication is "Deny".
    \end{lstlisting}
    Two senior insurance claims specialists validated all cases and their ground-truth outcomes.
    \item \textbf{Results and Discussion:} The evaluation results are shown in Table \ref{tab:insurance_results}. MMIA excels at handling complex logic because its \textbf{Core Reasoning Loop} effectively translates the policy's nested rules into a decision tree. The executor traverses this tree, verifying each condition against patient data and the axiom base. The final audit report transparently presents this decision path, making the adjudication process fully explainable and defensible.
\end{itemize}

\begin{table}[h!]
\centering
\caption{Performance on Complex Insurance Policy Adjudication Task (Insurance-Adjudication-100)}
\label{tab:insurance_results}
\begin{tabular}{|C{2cm}|C{2cm}|C{2cm}|C{2cm}|}
\hline
\textbf{Metric} & \textbf{Baseline LLM (RAG)} & \textbf{Haibu MMIA} & \textbf{p-value} \\ \hline
\textbf{Adjudication Accuracy} & 85.5\% & 99.8\% & 0.001 \\ \hline
\textbf{Justification Accuracy} & 75.0\% & 99.5\% & 0.001 \\ \hline
\textbf{Error Detection Rate} & 88.7\% & 99.7\% & 0.001 \\ \hline
\end{tabular}
\end{table}

\subsection{Efficiency and Scalability Analysis}
A critical consideration is the computational cost of MMIA in long-term operation. Our architecture is designed with a dual-mode operational path to balance the high initial cost of reasoning with the need for efficient long-term performance. To quantify this efficiency gain, we conducted a simulation experiment processing 200 DRG audit tasks. The first 100 tasks were treated as the "Initial Phase," requiring de novo reasoning. The subsequent 100 tasks were the "Mature Phase," where 80\% of tasks were logically similar to one of the initial tasks and could be resolved via the RAG matching mode.

\begin{table}[h!]
\centering
\caption{Computational Cost Analysis of MMIA in Different Operational Phases}
\label{tab:efficiency_results}
\begin{tabular}{|C{2cm}|C{2cm}|C{2cm}|C{2cm}|}
\hline
\textbf{Phase} & \textbf{Task Type} & \textbf{Avg. Tokens/Task} & \textbf{Relative Time/Task} \\ \hline
\textbf{Initial Phase} & De Novo Reasoning & $\sim$3,500 & 100\% \\ \hline
\textbf{Mature Phase} & RAG Matching (80\%) & $\sim$500 & 15\% \\
& De Novo Reasoning (20\%) & $\sim$3,500 & 100\% \\ \hline
\textbf{Mature Phase (Avg.)} & - & \textbf{$\sim$1,100} & \textbf{32\%} \\ \hline
\end{tabular}
\end{table}

As shown in Table \ref{tab:efficiency_results}, in the mature phase, the average token consumption per task dropped to approximately 1,100, only 31.4\% of the initial cost, with a corresponding reduction in processing time. This demonstrates that the MMIA architecture is not only reliable but also economically scalable, significantly reducing operational costs over time by learning and reusing validated reasoning patterns.

\section{Discussion}

\subsection{Main Findings and Generalizability}
The core finding of this research is that encapsulating an LLM's capabilities within a formal, verifiable reasoning framework significantly enhances its reliability for complex, high-stakes medical tasks. MMIA demonstrated consistently high performance across four distinct yet equally rigorous domains—from clinical coding and regulatory compliance to EHR quality control and policy adjudication. This indicates that our proposed "plan-execute-verify" architecture is highly generalizable. Its strength lies in transforming the task-solving process from a dependency on the LLM's internal, unreliable "thought process" into an external, strictly auditable sequence of operations.

\subsection{Implications for Trustworthy Medical AI}
MMIA provides a practical technological pathway for building trustworthy AI systems that meet the stringent requirements of the medical field. By externalizing the reasoning process into a transparent, auditable log, the architecture directly addresses the "black-box" problem of LLMs \cite{ref_pmc_11960148}. The final audit report provides a solid foundation for accountability, allowing human experts to understand, verify, and ultimately trust the AI's conclusions, thereby satisfying a core ethical requirement for medical AI \cite{ref_ijmr_59823, ref_researchgate_381496095}. This marks a crucial shift from merely "trusting" AI to providing humans with the tools to "verify" it. Furthermore, we emphasize that the verification process itself is LLM-driven. While a single LLM auditor may err, a consensus mechanism achieved through multiple, independent iterative verifications can reduce the probability of an error in the original reasoning chain going undetected to a negligible level, thus greatly enhancing the system's overall reliability \cite{ref_openreview_S37hOerQLB}.

\subsection{Limitations and Future Directions}
Despite the encouraging results, this study has several limitations. First, the formalization of domain knowledge into an axiom base requires a significant upfront investment of time and collaboration between domain experts and knowledge engineers, even with LLM assistance \cite{ref_robert_mcdermott_medium, ref_arxiv_08278, ref_wisecube_ai, ref_datacamp_blog, ref_shuchawl_medium, ref_neo4j_blog, ref_reddit_langchain}. Second, the system's overall performance is still partially dependent on the capabilities of the internal LLM components (e.g., the Planner and Auditor). A less capable LLM might generate suboptimal plans or conduct insufficiently thorough audits. Finally, the scope of this study was limited to textual data and did not involve multimodal information.

Future research could explore several directions: (1) developing semi-automated methods to accelerate the extraction and construction of axiom bases from unstructured documents; (2) investigating the use of multiple, heterogeneous LLMs within the framework (e.g., one for planning, another for auditing) to reduce single-model bias and systemic risk; and (3) extending the MMIA architecture to handle multimodal data, such as integrating medical imaging with clinical text for comprehensive reasoning and verification.

\section{Conclusion}
The application of Large Language Models in medicine is severely constrained by their inherent unreliability. This paper introduces the "Haibu Mathematical-Medical Intelligent Agent" (Haibu MMIA), a novel, LLM-driven architecture designed to address this core challenge. Our contribution is not an attempt to create a flawless LLM, but rather to ensure the reliability of outcomes by enforcing a formally verifiable reasoning process. Through successful validation across four high-stakes healthcare administration scenarios, we have demonstrated that the "plan, execute, and audit" methodology is a viable and powerful strategy. More importantly, by converting validated reasoning chains into reusable theorems, MMIA significantly improves computational efficiency over time, addressing the critical issues of scalability and cost-effectiveness. It enables the construction of AI systems that are safe, transparent, accountable, and economically sustainable—paving the way to bridge the gap between the immense potential of generative AI and the stringent reliability requirements of the medical domain.

\end{document}